\documentclass[sigconf,natbib=true,anonymous=false]{acmart}

\usepackage{amsmath}
\usepackage{amsthm,amsopn,bm}
\usepackage[noend]{algpseudocode}
\usepackage{array}
\usepackage{algorithm}
\usepackage{balance}
\usepackage{booktabs}
\usepackage{bigstrut}
\usepackage{booktabs}
\usepackage{blindtext}
\usepackage{caption}
\usepackage{color,soul}
\usepackage{enumitem}
\usepackage{footnote}
\usepackage{graphicx}
\usepackage{longtable}
\usepackage{multirow}
\usepackage{mathtools}
\usepackage{stmaryrd}
\usepackage{subcaption}
\usepackage{tabularx}
\usepackage{tabulary}
\usepackage{tikz}
\usepackage{url}
\usepackage{verbatim}
\usepackage{xcolor,colortbl}

\definecolor{green}{rgb}{0.1,0.1,0.1}

\definecolor{gitred}{HTML}{FDB8C0}
\definecolor{gitgreen}{HTML}{006400}
\definecolor{chocolate}{HTML}{D2691E}
\definecolor{maroon}{HTML}{800000}
\definecolor{indigo}{HTML}{4B0082}
\definecolor{green}{HTML}{008000}
\definecolor{orange}{HTML}{fc8d62}
\definecolor{purple}{HTML}{8da0cb}

\newcommand{\Fanswer}{$\mathrm{F1}_\text{ans}$~}

\newcommand{\Fcontro}{$\mathrm{F1}_\text{CONTRO}@e$~}
\newcommand{\FcontroOne}{$\mathrm{F1}_\text{CONTRO}@1$~}

\AtBeginDocument{%
  \providecommand\BibTeX{{%
    \normalfont B\kern-0.5em{\scshape i\kern-0.25em b}\kern-0.8em\TeX}}}

\copyrightyear{2023} 
\acmYear{2023} 
\setcopyright{acmlicensed}\acmConference[SIGIR '23]{Proceedings of the 46th International ACM SIGIR Conference on Research and Development in Information Retrieval}{July 23--27, 2023}{Taipei, Taiwan}
\acmBooktitle{Proceedings of the 46th International ACM SIGIR Conference on Research and Development in Information Retrieval (SIGIR '23), July 23--27, 2023, Taipei, Taiwan}
\acmPrice{15.00}
\acmDOI{10.1145/3539618.3591907}
\acmISBN{978-1-4503-9408-6/23/07}

\acmSubmissionID{res6645}

\begin{document}

\title{MythQA: Query-Based Large-Scale Check-Worthy Claim Detection through Multi-Answer Open-Domain Question Answering}

\author{Yang Bai}
\email{baiyang94@ufl.edu}
\affiliation{%
  \institution{The University of Florida}
  \city{Gainesville}
  \state{Florida}
  \country{USA}
  \postcode{32611}
}

\author{Anthony Colas}
\email{acolas1@ufl.edu	}
\affiliation{%
  \institution{The University of Florida}
  \city{Gainesville}
  \state{Florida}
  \country{USA}
  \postcode{32611}
}

\author{Daisy Zhe Wang}
\email{daisyw@ufl.edu}
\affiliation{%
   \institution{The University of Florida}
  \city{Gainesville}
  \state{Florida}
  \country{USA}
  \postcode{32611}}

\renewcommand{\shortauthors}{Yang Bai, Anthony Colas, \& Daisy Zhe Wang}

\begin{abstract}
Check-worthy claim detection aims at providing plausible misinformation to downstream fact-checking systems or human experts to check. This is a crucial step toward accelerating the fact-checking process. Many efforts have been put into how to identify check-worthy claims from a small scale of pre-collected claims, but how to efficiently detect check-worthy claims directly from a large-scale information source, such as Twitter, remains underexplored. To fill this gap, we introduce MythQA, a new multi-answer open-domain question answering(QA) task that involves contradictory stance mining for query-based large-scale check-worthy claim detection. The idea behind this is that contradictory claims are a strong indicator of misinformation that merits scrutiny by the appropriate authorities. To study this task, we construct TweetMythQA, an evaluation dataset containing 522 factoid multi-answer questions based on controversial topics. 
Each question is annotated with multiple answers. Moreover, we collect relevant tweets for each distinct answer, then classify them into three categories: "Supporting", "Refuting", and "Neutral". In total, we annotated 5.3K tweets. 
Contradictory evidence is collected for all answers in the dataset.  
Finally, we present a baseline system for MythQA and evaluate existing NLP models for each system component using the TweetMythQA dataset. We provide initial benchmarks and identify key challenges for future models to improve upon. 
Code and data are available at: \url{https://github.com/TonyBY/Myth-QA}
\end{abstract}





\maketitle

\section{Introduction}
\begin{figure}[!t]
\begin{center}
\scalebox{1.0}{
\begin{tabular}{@{}p{26em}}
\toprule 
\textbf{Question: What can spread COVID-19?} \\
\midrule
\textbf{Answer1: shoes}\\
\textbf{Supporting tweet}: Please stop running outdoors. Your running shoes could have coronavirus. There's a global pandemic. Stop the spread.\\

\textbf{Refuting tweet}: Fake news alert - "shoes carry virus" - No, infected people that spread virus wearing shoes do. Are you going to drink my muddy footprints? Madness in ignorance and stupidity abounds. I will exercise totally alone with no risk to anyone. Please understand\\
\midrule
\textbf{Answer2: swimming water}\\
\textbf{Supporting tweet}: Can Covid 19 virus spread through water? Seems so. Aisa hai toh will you wear masks while dancing in the Mumbai rains or while cooling off in a swimming pool? The latter can be dangerous, mind you.\\

\textbf{Refuting tweet}: Swimming Pool Water Unlikely to Spread Coronavirus But Facility Environments Need Careful Handling, Says Expert - Swimming World News\\
\bottomrule
\end{tabular}}
\end{center}
\caption{A TweetMythQA example. Note this entity question has 22 distinct answers annotated in the TweetMythQA. Two of the answers are listed here for demonstration purposes. MythQA aims to answer a factoid question with all distinct plausible answers and find contradictory stance evidence for each answer from a large corpus of tweets when available.}
\label{fig:TweetMythQA-example}
\end{figure}

\begin{figure}[t]
    \centering
    \captionsetup{justification=centering}
    \includegraphics[scale=0.55]{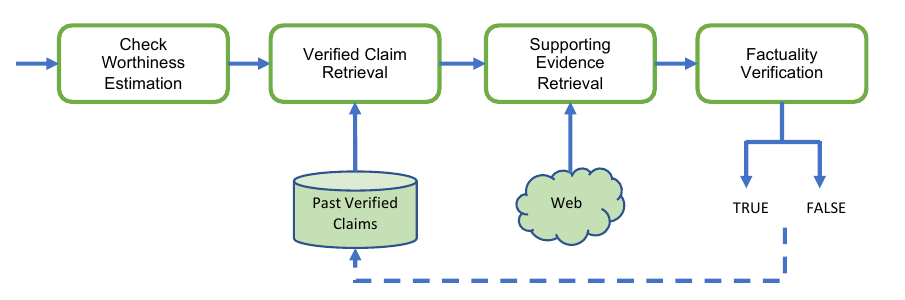}
    \caption{A traditional fact-checking pipeline.} 
    \label{fig:fact-pipeline}
\end{figure}

This paper proposes a new open-domain question answering (QA) task, MythQA, to tackle query-based large-scale check-worthy claim detection in an open-domain setting. Traditionally, the check-worthy claim detection module is used to determine which sentences in a given input text should be prioritized for fact-checking \citep{hassan2015quest, hassan2017toward}.
Figure~\ref{fig:fact-pipeline} shows the general steps of such a pipeline \citep{shaar-etal-2020-known}.

However, this setting is not efficient enough, especially when users are looking for distinct check-worthy claims on specific topics. It requires text snippets as input, which means that users still have to collect relevant passages of concerned topics through other methods for the system to check. Consequently, it is not suitable for detecting check-worthy claims on a large scale. What's more, the traditional setting does not provide evidence for the predictions, which reduces its interpretability and requires an extra step (e.g., supporting evidence retrieval) before the final step of fact-checking.

A key characteristic of open-domain question answering (QA) is its use of external information sources, such as Wikipedia, to answer users' factoid questions. Recently, multi-answer open-domain QA has been proposed by  \citep{Min2020AmbigQAAA}. One of its abilities is to help users find equally plausible distinct answers to users' queries. To introduce this ability in the check-worthy claim detection task, we propose MythQA(Query-based Large-scale Check-Worthy Claim Detection through Multi-Answer Open-domain Question Answering), the first open-domain QA-based large-scale distinct check-worthy claim detection framework which involves multi-answer open-domain question answering and contradictory stance mining. Specifically, the system must (1) find a set of distinct answers to the question, and (2) provide both supporting evidence and refuting evidence for each answer when available. The assumption is that any answer with contradictory stance evidence found may be misinformation that needs to be verified. New evaluation metrics are introduced in section~\ref{sec: task_setup}.

To support the study of this task, we construct a dataset called TweetMythQA with 522 multi-answer questions manually created. All questions are annotated with multiple answers. Moreover, we collect relevant tweets for each distinct answer, then classify them into three categories: "Supporting", "Refuting", and "Neutral". In total, we annotated 5.3K tweets. We make sure every answer is accompanied by contradictory evidence (both supporting and refuting evidence). Two types of questions are covered by TweetMythQA, according to  \citep{He2018DuReaderAC}: (1) factoid entity questions and (2) factoid yes/no questions. In the rest of the paper, we omit the term "factoid" when referring to the two types of questions. Details about data collection are provided in section~\ref{sec: Data Collection}.

We build the dataset using tweets because compared to other information sources such as government websites, Wikipedia, and news articles, Twitter contains more contradictory claims on many topics due to relatively weaker censorship. This leads to tweets spreading misinformation more rapidly, usually resulting in more severe consequences \citep{shu2017fake, Hossain2020COVIDLiesDC, Yang2020AnalysisAI, Alam2021FightingTC}. Furthermore, as far as we are aware, there is no open-domain QA dataset based on social media. Our dataset presents distinct challenges compared to existing open-domain QA systems. 

We provide benchmark results for existing zero-shot NLP models for this task. 
First, we examine the ability of information retrieval models to retrieve relevant tweets that include all distinct answers to a specific question. Second, we evaluate machine reading comprehension models on their ability to predict all distinct answers to the question given a set of relevant tweets. Third, following prior tasks  \citep{thorne-etal-2018-fact, Hossain2020COVIDLiesDC}, we evaluate NLI models on misinformation (a.k.a stance detection) by equating the class labels, Supporting, Refuting, and Neutral to Entailment, Contradiction, and Neutral, respectively. Finally, an end-to-end evaluation is applied over a newly proposed pipeline system for MythQA consisting of the above NLP modules. New evaluation metrics are proposed to take multiple answers and contradictory stance evidence into consideration. Our results show that there is significant room for future work on this task.

The main contributions of our work can be summarized as follows:
\begin{enumerate}
\item[(1)] We introduce MythQA, a new task that extends current multi-answer open-domain QA to include contradictory stance evidence mining for each distinct answer in order to detect check-worthy claims at a large scale.
\item[(2)] We construct TweetMythQA, a dataset with 522 multi-answer open-domain questions on multiple controversial topics; 5.3K relevant stance evidence in total is collected from Twitter for the answers.

\item[(3)] We introduce baseline systems for MythQA and evaluate existing NLP systems for each component of the system using the TweetMythQA dataset, providing initial benchmarks and identifying key challenges for future models to improve upon.
\end{enumerate}

\section{Related Works}
\subsection{Open-domain Question Answering}
Open-domain question answering(QA) aims to answer a factoid question in natural language by searching for evidence from a large-scale external information source such as Wikipedia~ \citep{voorhees1999trec,chen-etal-2017-reading}. 

Recently, many benchmarks have been contracted to help the community better understand this task, such as CuratedTREC \citep{Baudis2015ModelingOT}, WebQuestions \citep{Berant2013SemanticPO}, WikiMovies \citep{Miller2016KeyValueMN}, and Qusar-T \citep{Dhingra2017QuasarDF}. Several well-known machine reading comprehension(MRC) datasets were also curated to serve as open-domain QA benchmarks, such as OpenSQuAD \citep{chen-etal-2017-reading},  curated from SQuAD \citep{Rajpurkar2016SQuAD1Q}; OpenTriviaQA \citep{Lee2019LatentRF}, curated from TriviaQA  \citep{Joshi2017TriviaQAAL};  OpenNaturalQuestions \citep{Lee2019LatentRF}, curated from NaturalQuestions \citep{Kwiatkowski2019NaturalQA}. All of the benchmarks above use English Wikipedia as the external information source and all of them only require a single valid answer.

To study the ambiguity in open-domain questions,  \citep{Min2020AmbigQAAA} proposed the first multi-answer open-domain QA benchmark, AmbigQA. It requires a system to find all equally plausible answers to an ambiguous question and generate a more specific question for each distinct answer.

However, in open-domain questions, ambiguity is not the only cause for multiple answers. It is common to encounter contradictory answers and parallel answers in our daily lives, and in many cases, they are exactly what users seek. Figure~\ref{fig:TweetMythQA-example} shows examples of these cases, where "shoes" and "swimming water" are both plausible answers to the question "What can spread COVID-19?". By asking this question, users are likely seeking to find out all possible things that can be used as a COVID-19 medium, so they can avoid them all. Additionally, the "Supporting" and "Refuting" claims indicate that there are contradictory answers to questions such as "Can shoes spread COVID-19?" and "Can swimming water spread COVID-19?". These contradictory answers are valuable for users and fact-checking organizations to be aware of the potential risks. Hence, one of the main motivations for this study is to explore the underexplored challenges of open-domain quality assurance that arise when answers are contradictory and parallel. To our knowledge, MythQA is the first study of its kind.

In contrast to previous works that used Wikipedia as their external information source, MythQA focuses on social media (Twitter).  Compared to Wikipedia, social media has a significantly different data distribution as a result of the noise and informal nature of the text, and diverse (contradictory and parallel) claims about controversial topics. The features of social media pose new challenges for existing open-domain QA systems, but they provide a perfect corpus for MythQA, which aims to find check-worthy claims. 

Lastly, MythQA is the first open-domain multi-answer QA benchmark with an emphasis on yes/no questions. AmbigNQ (the dataset proposed along with AmbigQA), in total, has only five yes/no questions in their development and training set, but none are multi-answer questions. TweetMythQA, on the other hand, contains 408 yes/no questions that are all multiple-answer questions.

\subsection{Check-worthy Claim Detection}
Check-worthy claim detection(CWCD) aims at predicting which sentences in an input text should be prioritized for fact-checking.
The first work targeting the detection of check-worthy claims was the ClaimBuster system proposed by  \citep{hassan2015quest, hassan2017toward}. The model is trained on manually annotated data where each sentence was marked as check-worthy factual, non-factual, or unimportant factual. The data consisted of transcripts of historical US election debates. Later, a larger version of the ClaimBuster dataset was published \citep{arslan2020benchmark}. ClaimRank \citep{Jaradat2018ClaimRankDC} is another important system for detecting check-worthy claims. It extended ClaimBuster to a new language(Arabic) and paid special attention to the context of each sentence. Other well-known works focused on political debates including  \citep{Atanasova2018OverviewOT, Atanasova2019OverviewOT,
Gencheva2017ACA,
Patwari2017TATHYAAM}.

Recently, new domains are introduced for check-worthy claim detection, such as social media  \citep{Shaar2020OverviewOC, nakov2021overview}, news articles \citep{alhindi2021fact}, Wikipedia \citep{wright2020claim} and COVID-19 \citep{Alam2021FightingTC}. 

However, all previous works have focused on the setting of classification or ranking over sentences within a given text. As an alternative, MythQA aims to retrieve check-worthy claims directly from a large external information source (Twitter) based on users' queries.

\section{Task: MythQA}\label{sec: task_setup}

\begin{figure*}[t]
  \begin{subfigure}[t]{0.49\textwidth}
    \centering
    \includegraphics[width=\textwidth]{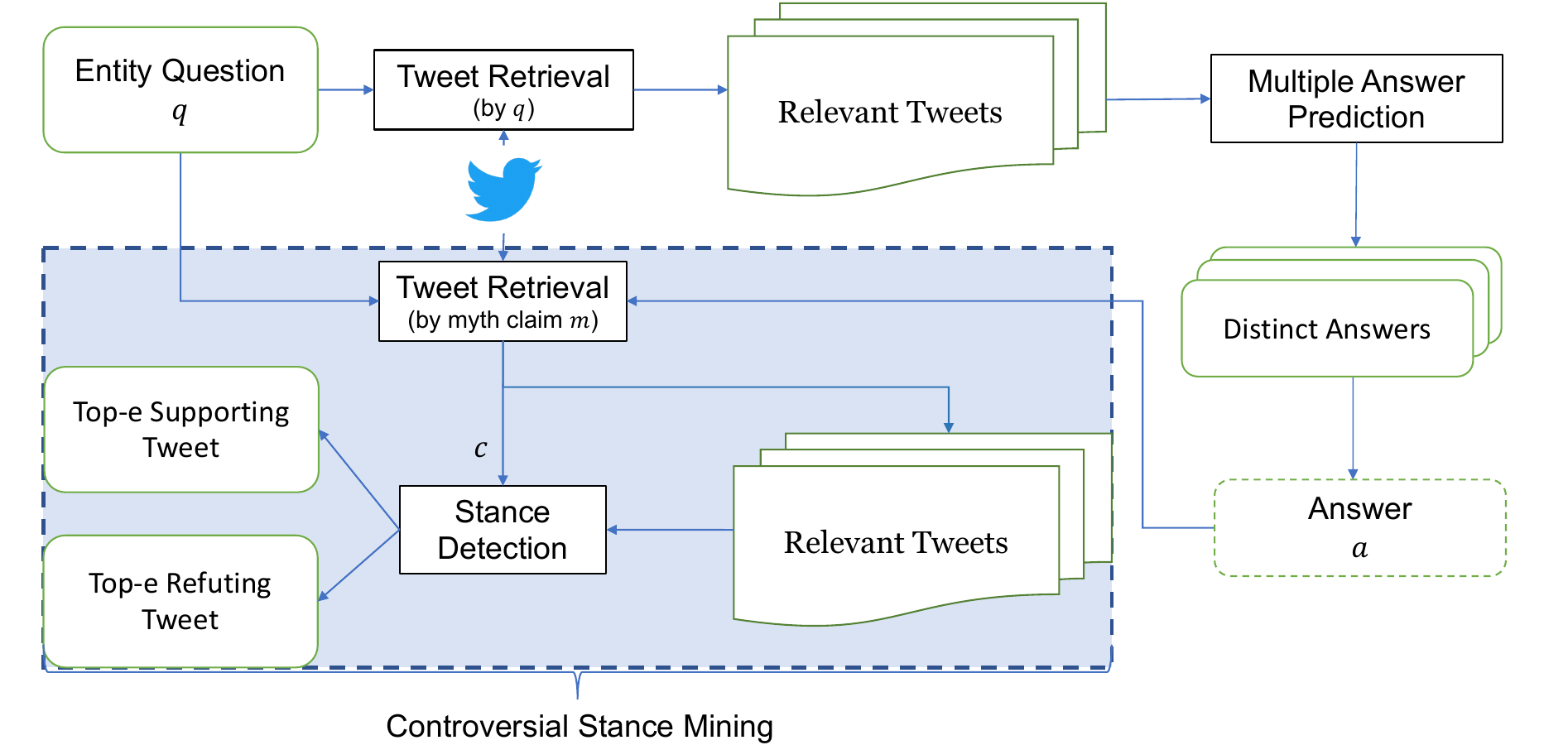}
    \caption{MythQA pipeline for entity questions. A claim $m$ is composed of a question $q$ and a distinct answer $a$.}
    \label{fig:1a}
  \end{subfigure}\hfill
  \begin{subfigure}[t]{0.49\textwidth}
    \centering
    \includegraphics[width=\textwidth]{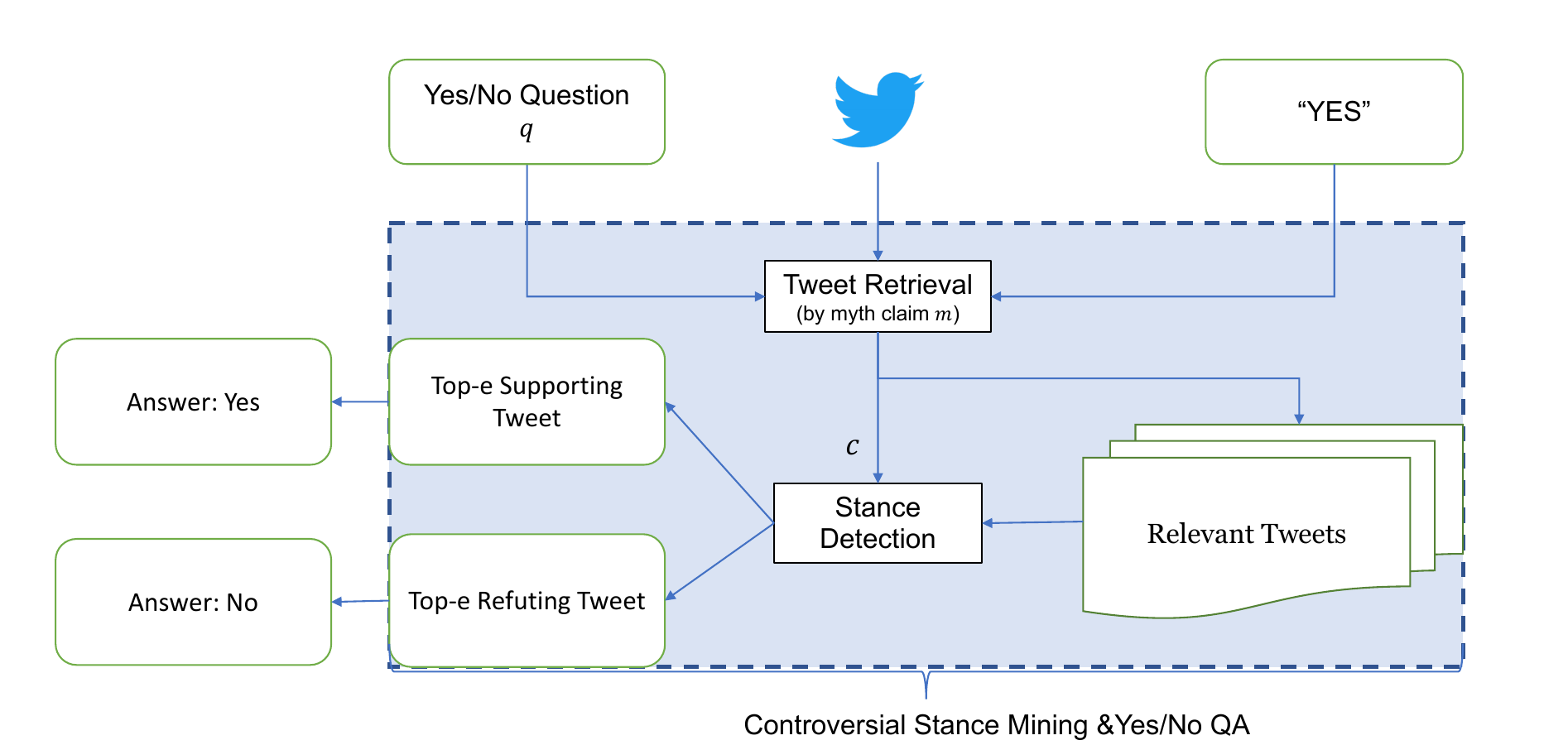}
    \caption{MythQA pipeline for yes/no questions. A claim $m$ is composed of a question $q$ and a positive answer "YES". 
    } 
    \label{fig:1b}
  \end{subfigure}

\caption{MythQA pipeline.} \label{fig:MythQA pipeline.}
\end{figure*}

\subsection{Task Setup}
In this section, we first describe the MythQA task setup for entity questions and yes/no questions, respectively. Then, we describe the task setups for each subtask individually.

For entity questions, given a question $q$, the task is to predict a set of (answer, supporting evidence, refuting evidence) tuples: $\{(a_{1}, \mathcal{S}_{1}, \mathcal{R}_{1}), (a_{2}, \mathcal{S}_{2}, \mathcal{R}_{2}), ..., (a_{n}, \mathcal{S}_{n}, \mathcal{R}_{n})\}$, where each $a_{i}$ is a plausible answer to $q$, each $\mathcal{S}_{i}$ is a set of supporting evidence tweet of $a_{i}$, and each $\mathcal{R}_{i}$ is a set of refuting evidence tweet of $a_{i}$. Note that to be more practical and increase the chance of finding gold evidence, we predict a set of supporting/refuting evidence instead of a single one.

For yes/no questions, since the contradictory evidence of answers "YES" and "NO" are symmetric, hence, a system is only required to find supporting evidence for each answer, respectfully. In particular, given a question $q$, predict a set of (answer, supporting evidence) pairs: $\{(a_{1}, \mathcal{S}_{1}), (a_{2}, \mathcal{R}_{2})\}$.\\

\noindent{\textbf{Tweet retrieval}} \leavevmode \\
Given a question $q$ or a claim $c$, output top-$k$ relevant tweets, $T = \{t_{1}, t_{2}, ..., t_{k}\}$ from a large tweet corpus $C$, where $k$ is a given hyperparameter.\\

\noindent{\textbf{Stance Detection}}  \leavevmode \\
Given an $(c, t)$ pair, predict their stance relation from Supporting, Refuting, and Neutral.\\

\noindent{\textbf{Multiple Answer Prediction}} \leavevmode \\
Given a question $q$ and the top-$k$ relevant tweets $T$, output a set of distinct plausible answers $A = \{a_{1}, a_{2}, ..., a_{n}\}$, where $n$ is unknown.\\

\noindent{\textbf{Contradictory Stance Mining}}  \leavevmode \\
Given a claim $c$, return the top-$e$ supporting tweet evidence, and the top-$e$ refuting tweet evidence from a large tweet corpus $C$ when available, where $e$ is a given hyperparameter. This task could be thought of as a combination of relevant claim retrieval and stance detection.

A claim $c$ can be constructed automatically as simply as by concatenating a distinct answer and its corresponding question, e.g, "What can spread COVID-19? Answer is swim water.", where " Answer is " serves as a connection string.

\subsection{Evaluation Metrics}\label{subsec: Evaluation Metrics}
 To evaluate model performance, we present several ways to compare a model prediction with $m$ (answer, supporting evidence, refuting evidence) tuples $(a_1, \mathcal{S}_1, \mathcal{R}_1), \dots, (a_m, \mathcal{S}_m, \mathcal{R}_m)$
with a gold reference set with $n$ tuples
$(\bar{a}_1, \bar{\mathcal{S}}_1, \bar{\mathcal{R}}_1), \dots, (\bar{a}_n, \bar{\mathcal{S}}_n, \bar{\mathcal{R}}_n)$, where $size(\mathcal{S}_i) = size(\mathcal{R}_i) = e$. 

We assign each predicted (answer, supporting evidence, refuting evidence) tuple $(a_i, \mathcal{S}_i, \mathcal{R}_i)$ a \textit{supporting evidence match score} $se_i$  and a \textit{refuting evidence match score} $re_i$ based on if any gold evidence is included. We use exact match when comparing the predicted answer/evidence and the gold answer/evidence.

\begin{equation}
se_i=\max_{\substack{1 \leq j \leq n}} \mathbb{I}[a_i = \bar{a}_j] f(\mathcal{S}_i, \bar{\mathcal{S}}_j)
\end{equation}

\begin{equation}
re_i=\max_{\substack{1 \leq j \leq n}} \mathbb{I}[a_i = \bar{a}_j] f(\mathcal{R}_i, \bar{\mathcal{R}}_j)
\end{equation}

where $f$ is defined as:

\begin{equation}\label{eq:f}
f(E, \bar{E})=\mathbb{I}[E \cap \bar{E} \neq \emptyset]
\end{equation}

where $E$ is a set of evidence text.

The correctness score $c_i$ of answer $a_i$ is the average of $se_i$ and $re_i$:
\begin{equation}
c_i = (se_i + re_i)/2
\end{equation}

Note that, for a yes/no question, finding correct supporting evidence for the NO answer is equivalent to finding correct refuting evidence for the YES answer, and vice versa, hence, $c_i$ could be simplified to:

\begin{equation}
c_i = se_i
\end{equation}

We calculate F1 treating the $c_i$ as measures of correctness: 

\begin{equation}
\mathrm{prec}_{f} = \frac{\sum_i c_i}{m}
\end{equation}

\begin{equation}
\mathrm{rec}_{f} = \frac{\sum_i c_i}{n}
\end{equation}

\begin{equation}
\mathrm{F1}_{f} = \frac{2 \times  \mathrm{prec}_{f} \times \mathrm{rec}_{f}}{\mathrm{prec}_{f} + \mathrm{rec}_{f}}
\end{equation}

We consider two choices for $\mathrm{F1}_{f}$: \textbf{\Fanswer}, and \textbf{\Fcontro}. \textbf{\Fanswer} is the F1 score on answers only, where $f$ always yields 1. \\
\textbf{\Fcontro} further checks if any gold evidence is included in the top-e evidence predictions, where $f$ is defined in equation~(\ref{eq:f}).\\

To evaluate a retriever's result with respect to the number of distinct answers found in the top-k retrieval, we propose
$MHits@k$:

\begin{equation}
MHits@k = Hits@k \times \frac{\sum_i \mathbb{I}[\bar s_i \in T \ \lor \  \bar r_i \in T]}{n}
\end{equation}

For failing to retrieve relevant evidence for all distinct answers to the question, the retriever will be penalized.

\begin{table}[t]
    \centering
    \setlength{\tabcolsep}{6.5pt} 
    \renewcommand{\arraystretch}{1.2} 
    \begin{tabular}{lccc}
        \toprule
            \textbf{Q Type} & \textbf{\# Q} 
              & \textbf{\# Avg.SupEve/Ans} & \textbf{\# Avg.RefEve/Ans} \\
        \midrule
            Entity & 114 
            & 2.89 & 2.47  \\
            Yes/No  & 408 
            & 2.92 & 2.49  \\
            Overall  & 522 
            & 2.91 & 2.49 \\
        \bottomrule
    \end{tabular}
    \caption{
	    Data statistics. In TweetMythQA, all questions have multiple answers. All answers have multiple pieces of contradictory evidence.
	}\label{tab:general-data-statistics}
\end{table}

\begin{table}[t]
    \centering
    \setlength{\tabcolsep}{6.5pt} 
    \renewcommand{\arraystretch}{1} 
    \begin{tabular}{cccccc}
        \toprule
            & \multicolumn{4}{c}{\textbf{\# Ans / Q}}\\\cmidrule(lr){2-5}
            \textbf{Question Type} & 1 & 2 & 3 & 4+ & \# Avg.Ans/Q\\
        \midrule
            Entity & 0  & 41 & 35  & 38 & 3.43\\
            Yes/No  & 0   & 408 & 0  & 0  & 2\\
        \bottomrule
    \end{tabular}
    \caption{
	    The number of answers per question. 
	}\label{tab:answer-statistics}
\end{table}

\begin{table}[t]
    \centering
    \setlength{\tabcolsep}{6.5pt} 
    \begin{tabular}{ccc}
        \toprule
            \textbf{Stance Label} 
            & \textbf{Count}
            & \textbf{Percentage} \\
        \midrule
            Supporting & 2318 &  43.88\%  \\
            Refuting  & 1980 & 37.49\%    \\
            Neutral & 984 & 18.63\% \\
        \bottomrule
    \end{tabular}
    \caption{
	    Distribution of evidence by stance. 
	}\label{tab:stance-statistics}
\end{table}

\section{Data Collection}\label{sec: Data Collection}
\begin{figure*}[t]
    \centering
    \captionsetup{justification=centering}
    \includegraphics[scale=1.15, angle =0]{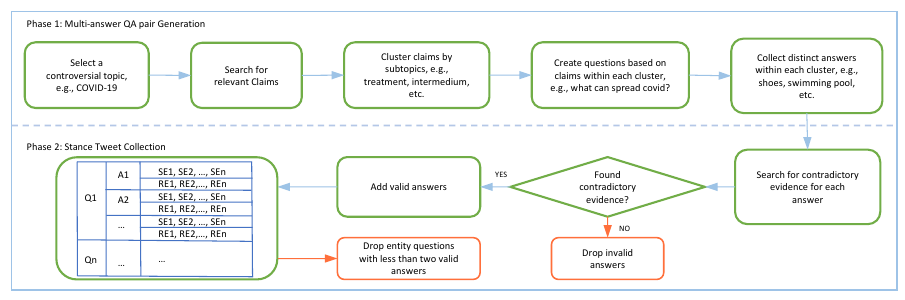}
    \caption{Annotation process of TweetMythQA. The terms Q, A, SE, and RE refer to the question, answer, supporting evidence, and refuting evidence, respectively.} 
    \label{fig:annotation_process}
\end{figure*}

We construct TweetMythQA examples in two phases: multi-answer QA pair generation and stance evidence collection. An overview of the annotation process is shown in Figure~\ref{fig:annotation_process}. We build a large external information corpus using cleaned relevant tweets collected during annotation using the Twitter API. We ensure that (1) all the tweets in the corpus are relevant to the topics of the questions in TweetMythQA; (2) all the tweet ids in TweetMythQA are included in the Corpus; (3) retweets and duplicate tweets are removed. In total, around 200K English tweets were collected as an external information source. For training and quality control, high-quality annotators are carefully selected and hired from three master students in the computer science department. Annotations are sampled and checked by co-authors during the process to ensure they are high-quality. 

\subsection{Annotation Process}

\noindent{\textbf{Multi-Answer QA Pair Generation}} \leavevmode \\
The process of creating a question in TweetMythQA involves three major steps: selecting a controversial topic, searching for relevant claims, and creating the question. In particular, to create an entity question, annotators must summarize claims from the same topic, while yes/no questions are created by directly paraphrasing claims from the selected topics into verification questions\footnote{Noted that the claims used to create entity questions can also be used to create yes/no questions, resulting in the fact that many yes/no questions in our dataset have the same format as entity questions. For example, "What can spread COVID-19?" and "Can shoes spread COVID-19?". We allow such overlap in the dataset because we evaluate entity questions and yes/no questions separately.}. 

The social media surrounding controversial topics tends to contain many contradictory claims. Thus, when generating questions, a controversial topic (e.g., COVID-19) is selected from a list of controversial topics on Wikipedia\footnote{\url{https://en.wikipedia.org/wiki/Wikipedia:List_of_controversial_issues}}.

Then, relevant claims are gathered through myth-busting and fact-checking websites, such as WHO\footnote{\url{https://www.who.int/emergencies/diseases/novel-coronavirus-2019/advice-for-public/myth-busters}}, FEMA\footnote{\url{https://www.fema.gov/disasters/coronavirus/rumor-control}}, 
Wikipedia\footnote{\url{https://en.wikipedia.org/wiki/COVID-19_misinformation}}, 
NewsGuard\footnote{\url{https://www.newsguardtech.com}}, 
Google Fact Check Tools\footnote{\url{https://toolbox.google.com/factcheck/explorer}}, 
EUvsDisinfo\footnote{\url{https://euvsdisinfo.eu/disinformation-cases/}}, 
Polygraph.info\footnote{\url{https://www.polygraph.info/}}, PolitFact\footnote{\url{https://www.politifact.com/}}, etc

After retrieving the claims, they are manually clustered based on different aspects/subtopics of the controversial topic, such as medium, prevention, source, etc.

By summarizing the claims in each cluster, the annotator formulates an entity question that can be answered by any of the distinct claims within the cluster. For example, based on claims of the same subtopic cluster: "The likelihood of shoes spreading COVID-19 is very low", "Water or swimming does not transmit the COVID-19 virus", etc. A question such as: "What can spread COVID-19?" could be created.

Finally, short answers are derived from the claims, for example, shoes, swimming water, etc. Following  \citep{Lee2019LatentRF}, all annotated answers are shorter than 5 tokens.\\

\noindent{\textbf{Stance Tweets Collection}}\leavevmode \\
Each distinct answer and corresponding question make up a distinct claim. In the second phase, we collect contradictory stance evidence for every distinct claim through the Twitter API. Annotators are rewarded for finding both supporting and refuting evidence for each answer.

We observe that the results directly returned by the Twitter API are very noisy, which makes it hard to find relevant tweets that can be used as stance evidence for most of the claims collected from the first-phase annotation, let alone finding contradictory evidence.  

To reduce the annotation difficulty, we developed a heuristic tool that can help annotators search through the Twitter API and suggests the top 100 potentially controversial tweets to annotators. In particular, it can automatically construct multiple queries for each claim based on a template and a list of alias of the topic entity given by the annotators. For example, given a template: "shoes can spread TOPIC\_ENTITY", and a list of alias of the TOPIC\_ENTITY, e.g.,  \emph{COVID-19, covid, Coronavirus, Coronavirus 2, SARS-CoV-2, novel coronavirus-2019, Wuhan virus, etc}. Queries of the same template with different alias of the TOPIC\_ENTITY will be constructed and sent to the Twitter API. We find this practice can significantly increase the chance of finding contradictory stance evidence. 

For the raw tweets returned by the Twitter API, the tool does a series of data cleaning such as: dropping retweets and tweets with similar content, etc. After that, a state-of-the-art pretrained dense passage ranker, DPR (Reference), is applied to retrieve the top-1000 tweets based on the semantic similarity between the query and the cleaned tweets. Then the top-1000 tweets are clustered in 5 clusters using the K-means algorithm based on their embeddings generated by the indexer that is used by the DPR. Finally, 100 tweets consisting of the top-20 tweets of each cluster are returned as the final set for the annotators to check.

\subsection{Dataset Statistics}

\begin{figure*}[t]
    \centering
    \captionsetup{justification=centering}
    \includegraphics[scale=0.37, angle =0]{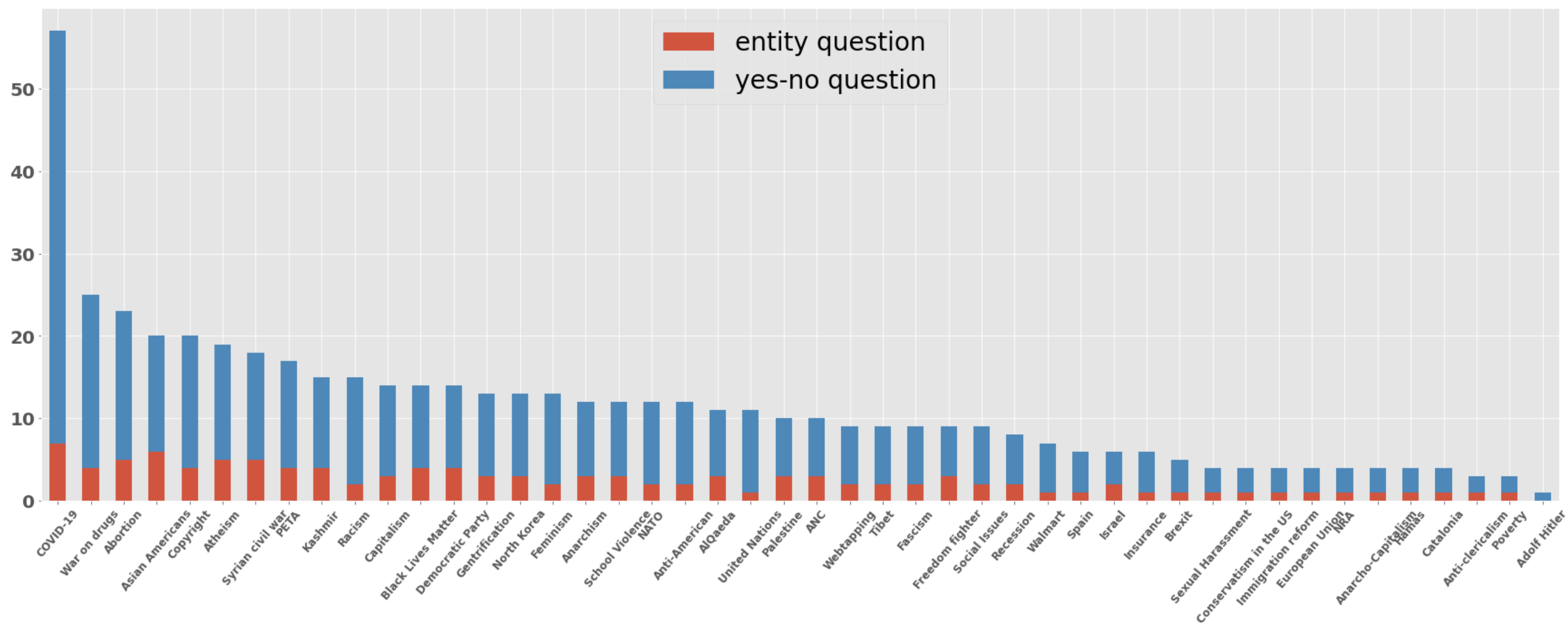}
    \caption{Distribution of topics and question types in TweetMythQA.} 
    \label{fig:question_distribution}
\end{figure*}

In TweetMythQA, two types of factoid questions are annotated: entity questions and yes/no questions. Table~\ref{tab:general-data-statistics} shows that in TweetMythQA, all questions are multi-answer questions, and each answer is backed up with multiple pieces of contradictory evidence(both supporting and refuting evidence). Table~\ref{tab:answer-statistics} gives more detailed statistics about the number of distinct answers per question. Each entity question has an average of 3.5 distinct answers, and all yes/no questions have an average of two distinct responses. Table~\ref{tab:stance-statistics} shows the distribution of stance evidence. In total, there are 44 topics covered by TweetMythQA. 

In total, there are 44 topics covered by TweetMythQA. 
Figure~\ref{fig:question_distribution} shows the distribution of each type of question among topics in TweetMythQA. By the category of the controversial topic list on Wikipedia, besides COVID-19 is from the "Science, biology, and health" section, all the rest topics in the TweetMythQA are selected from the section "Politics and economics" due to limited annotation time. More topics in other sections will be covered in our future work.

\subsection{Quality Control}
Due to the significant portion of search and summary procedures involved, and often the requirement of strong background knowledge for some topics, this annotation task is significantly more difficult than constructing other single-answer, single-stance-evidence QA datasets, and normal stance detection datasets. We sample and check annotations frequently during the annotation process in order to ensure quality. In addition, we use handy communication tools such as Slack\footnote{\url{https://slack.com/}} and Zoom\footnote{\url{https://zoom.us/}} to answer any technical issues as soon as possible. Prior to annotating the remaining examples, any disagreements were discussed and resolved.

Furthermore, we ensure that each question has multiple answers, and each answer has both supporting and refuting evidence. As a result, there is a high annotation rejection rate, which results in a high-quality dataset at a high cost.

\subsection{More Details on Annotation Work}
The three annotators are all first-year graduate students in the computer science department. All of them are male. 

Annotators are paid 13 dollars/hour with certain expectations and rewards for different phases of annotation. 

Phase-one annotations require each annotator to complete four entity questions per hour and provide at least two distinct answers to each question.  More distinct answers to each entity question are encouraged. There will be a bonus of \$3.25 for every additional four distinct answers.  

Phase-two annotation requires each annotator to collect 120 pieces of supporting/refuting evidence per hour for the answers collected in the last annotation phase. Since it is difficult to find both supporting and refuting evidence, annotators are rewarded with the full hourly payment if they can find two pairs of non-overlapping supporting-refuting evidence for an answer, i.e., two distinct supporting evidence and two distinct refuting evidence.

\section{Performance of Benchmark Models}
In real life, claims generate and evolve rapidly. Annotating a dataset large enough for training would be very expensive and would never keep up with the speed at which data changes in the field of check-worthy claim detection (CWCD). Thus, it is desirable that CWCD systems be data-efficient, i.e., trained with little or no supervision.

In this section, we investigate various zero-shot baselines for different sub-tasks within our MythQA task outlined in Section~\ref{sec: task_setup}. We test a diverse set of baselines in an evaluation-only setting for the sub-tasks below. 

\subsection{Tweet retrieval}
We evaluate a sparse retrieval approach and a dense retrieval approach for the tweet retrieval sub-task. Specifically, for the sparse retrieval approach, we choose BM25  \citep{Robertson2009ThePR}, a famous traditional scoring algorithm using bag-of-words representations. For the dense retrieval approach, we choose DPR\footnote{\url{https://huggingface.co/facebook/dpr-question_encoder-multiset-base}}\footnote{\url{https://huggingface.co/facebook/dpr-ctx_encoder-multiset-base}} \citep{Karpukhin2020DensePR}, a state-of-the-art on NQ-open, which does nearest-neighbor search on transformer-encoded representations. We use the Pyserini IR toolkit \citep{Lin2021PyseriniAP} to implement both BM25 and DPR tweet retrievers.

\noindent\textbf{Results}~~~~ We present the performance of two tweet retrievers in Table~\ref{tab:TweetRetrievalResults}. For entity questions and yes/no questions, we see that BM25 performs better than pretrained DPR. This advantage is more significant for yes/no questions especially when the retrieval number is small. Besides, we find that $MHit$ scores are clearly lower than the $Hit$ scores for the entity questions. This indicates the challenges in multi-answer retrieval, i.e., maximizing the coverage of relevant evidence(supporting and refuting evidence) of distinct answers in the top-k retrievals.

\begin{table*}[htbp]
    \centering
    \setlength{\tabcolsep}{12pt} 
    \renewcommand{\arraystretch}{1.2} 
    \begin{tabular*}{\textwidth}{lccccccccc}
        \toprule
            & \multicolumn{4}{c}{\textbf{BM25}} & \multicolumn{4}{c}{\textbf{DPR}}\\\cmidrule(lr){2-5}\cmidrule(lr){6-9}
            \textbf{Question Type} & MH@100 & H@100 & MH@1K & H@1K & MH@100 & H@100 & MH@1K & H@1K\\
        \midrule
            Entity  & 79.67 & 92.98 & \textbf{96.13} & 99.12
                    & 70.29 & 88.60 & \textbf{92.90} & 98.25\\
                     
            Yes/No  & 97.06 & 97.06 & \textbf{99.26} & 99.26 
                    & 80.88 & 80.88 & \textbf{95.10} & 95.10\\
            
            Overall & 93.26 & 96.17 & \textbf{98.58} & 99.23 
                    & 78.57 & 82.57 & \textbf{94.62} & 95.79\\
        \bottomrule
    \end{tabular*}
    \caption{
	    Tweet Retrieval Performance. We present evaluation results for relevant tweet retrieval (i.e., tweets that either support or refute one or more distinct answers). $MH@k$ is the abbreviation of $MHit@k$, a new metric that we introduce in section~\ref{subsec: Evaluation Metrics}. It is important to note that for yes/no questions, $Hit@k$ is equivalent to $MHit@k$ because it either finds relevant tweets for all answers ("YES" and "NO") or for no answers.
	}\label{tab:TweetRetrievalResults}
\end{table*}

\begin{table*}[t]
    \centering
    \setlength{\tabcolsep}{8pt} 
    \renewcommand{\arraystretch}{1} 
    \begin{tabular*}{\textwidth}{lcccccccccccc}
        \toprule
            & \multicolumn{3}{c}{\textbf{Macro Avg.}} &   \multicolumn{3}{c}{\textbf{Supporting}} & \multicolumn{3}{c}{\textbf{Refuting}} & \multicolumn{3}{c}{\textbf{Neutral}}\\\cmidrule(lr){2-4}\cmidrule(lr){5-7}\cmidrule(lr){8-10}\cmidrule(lr){11-13}
            \textbf{Model} & 
            P & R & F & 
            P & R & F & 
            P & R & F & 
            P & R & F\\
        \midrule
            BERT-large      & 23.95 & 35.47 & 27.63
                            & 52.68 & 76.79 & 62.49
                            & 3.70 & 0.05 & 0.10
                            & 15.47 & \textbf{29.58} & \textbf{20.31}\\
                            
            ALBERT-large    & 25.95 & 35.49 & 26.96
                            & 52.43 & \textbf{90.38} & 66.36
                            & 12.82 & 0.25 & 0.50
                            & 12.59 & 15.83 & 14.03\\
                            
            XLNet-large     & 30.20 & 33.05 & 27.50
                            & 49.26 & 81.67 & 61.45
                            & 30.40 & 5.80 & 9.75
                            & 10.94 & 11.67 & 11.29\\
                            
            BART-large      & 41.90 & 43.58 & 39.67
                            & 52.87 & 73.94 & 61.65
                            & 56.16 & 56.16 & 56.16
                            & 16.67 & 0.63 & 1.20\\
                            
            RoBERTa-large   & 48.21 & 47.62 & 43.90
                            & 54.98 & 85.76 & 67.00
                            & 70.24 & 55.75 & 62.16
                            & 19.40 & 1.35 & 2.53\\
            
            DeBERTa-large   & \textbf{50.12} & \textbf{48.90} & \textbf{44.74}
                            & \textbf{55.83} & 89.00 & \textbf{68.62}
                            & \textbf{73.85} & \textbf{57.08} & \textbf{64.39}
                            & \textbf{20.69} & 0.63 & 1.21\\
        \bottomrule
    \end{tabular*}
    \caption{
	    Stance Detection Performance. We present evaluation results for classifying claim-tweet pairs into Supporting, Refuting, and Neutral classes. The precision (P), recall (R), and F1-score (F1) are presented for each class, as well as macro averaged values. All models are pretrained on MNLI.
	}\label{tab:StanceDetectionResults}
\end{table*}

\subsection{Stance Detection}
Following \citep{Yin2019BenchmarkingZT, Hossain2020COVIDLiesDC}, we leverage NLI models for the zero-shot stance detection task. In particular, We evaluate existing NLI models: BERT-large \citep{Devlin2019BERTPO}\footnote{\url{https://huggingface.co/madlag/bert-large-uncased-mnli}}, ALBERT-large \cite{Lan2020ALBERTAL}\footnote{\url{https://huggingface.co/anirudh21/albert-large-v2-finetuned-mnli}}, XLNet-large \cite{Yang2019XLNetGA}\footnote{\url{https://huggingface.co/ynie/xlnet-large-cased-snli_mnli_fever_anli_R1_R2_R3-nli}},  BART-large \citep{Lewis2020BARTDS}\footnote{\url{https://huggingface.co/facebook/bart-large-mnli}}, Roberta-large \citep{Liu2019RoBERTaAR}\footnote{\url{https://huggingface.co/roberta-large-mnli}}, and DeBERTa \citep{He2021DeBERTaDB}\footnote{\url{https://huggingface.co/microsoft/deberta-large-mnli}}. All models are pretrained over MNLI \citep{Williams2018ABC}. We construct our stance detection evaluation dataset, TweetMythSD, which consists of claim-evidence pairs in the TweetMythQA. Each claim is composed of a distinct answer and the corresponding question as described in section~\ref{sec: task_setup}. Table~\ref{tab:stance-statistics} shows statistics of the TweetMythSD.

\noindent\textbf{Results}:~~~~ The results of the stance detection are shown in Table~\ref{tab:StanceDetectionResults}. DeBERTa-large achieves the highest F1 for the Supporting and Refuting class, as well as the highest macro-averaged Procession, Recall, and F1 for all the pretrained models we evaluated. On the other hand, most models do not perform well on the Neutral class.  In addition, we observe that Refuting and Neutral results differ greatly while Supporting results are relatively stable. The reason may be that, in our dataset, there is a greater data shift in refuting and neutral examples than in supporting examples. Different results between the models can be attributed to differing generalization abilities for each stance class.

\subsection{Multiple Answers Prediction}\label{subsec: Multiple Answers Prediction}
Due to the difference between the nature of entity questions and yes/no questions, we propose a specific pipeline for each of the two types of questions, as shown in Figure~\ref{fig:MythQA pipeline.}. 

\noindent\textbf{Entity Questions:} To predict distinct answers to entity questions, we use a pretrained machine reading comprehension(MRC) model. In particular, we evaluate an extractive MRC model DPR reader\footnote{\url{https://huggingface.co/facebook/dpr-reader-single-nq-base}} \citep{Karpukhin2020DensePR} pretrained on Natural Questions \citep{Kwiatkowski2019NaturalQA}, and a generative MRC model T5\footnote{\url{https://huggingface.co/valhalla/t5-base-qa-qg-hl}}  \citep{Raffel2020ExploringTL} pretrained on SQuAD \citep{Rajpurkar2016SQuAD1Q}. 

Neither the original DPR reader nor the T5 reader was designed to predict multiple answers. For this task, we propose a simple but efficient method for predicting multiple answers. Rather than train a model to decide the number of answers, $n$, to predict \citep{Min2020AmbigQAAA}, we make $n$ a hyperparameter, i.e., an MRC model only needs to find the top-$n$ distinct answers from the top-$k$ retrieved tweets, where $k>>n$. For each tweet, the MRC model only needs to predict one answer. Candidate answers are ranked based on a weighted average of tweet retrieval scores from the retriever model and answer span scores from the MRC model.\footnote{For generative MRC model such as T5, we only use the tweet retrieval score to rank the answer.}. A normalization process will also be applied to the answers. Answers that are duplicated will be removed.  By doing so, we can make direct use of MRC models pretrained on other large datasets of single-answer MRCs.
 
\noindent\textbf{Yes/No Questions:} Rather than using a machine reading comprehension (MRC) model to predict different yes/no answers, we directly do contradictory stance mining for the positive claims. Each positive claim is composed of a yes/no question and the "YES" answer as described in section~\ref{sec: task_setup}. Findings of supporting and refuting evidence of the positive claim, indicate the answers "YES" and "NO" to the yes/no question respectively. If no relevant evidence is found, output "NOT SURE". On the basis of the logits of the stance detection model, the top-ranked evidence of "YES" and "NO" answers is selected. 

Both intrinsic and extrinsic evaluations are done for entity and yes/no questions.

\noindent\textbf{Intrinsic}: The prediction is conditioned on a question and relevant tweets that are annotated for the question. 

\noindent\textbf{Extrinsic}: The prediction is conditioned on a question and relevant tweets retrieved by the tweet retriever from the whole corpus. \\

\begin{table}[t]
    \centering
    \setlength{\tabcolsep}{4pt} 
    \renewcommand{\arraystretch}{1.2} 
    \begin{tabular}{lccccccc}
        \toprule
            & \multicolumn{3}{c}{\textbf{Intrinsic (\Fanswer)}} & \multicolumn{3}{c}{\textbf{Extrinsic (\Fanswer)}}\\\cmidrule(lr){2-4}\cmidrule(lr){5-7}
            \textbf{Models} & m=1 & m=5 & m=10 & m=1 & m=5 & m=10\\
        \midrule
        &&&k = 10\\
        \midrule
            DPR Reader  & 16.10 & 26.56 & 26.94 
                        & 14.59 & 20.68 & 19.84\\
                     
            T5          & 16.45 & 25.45 & 27.06
                        & 12.17 & 20.20 & 19.69\\
        \midrule
        &&&k = 100\\
        \midrule
            DPR Reader  & 16.51 & 27.88 & 29.69 
                        & 15.10 & \textbf{20.83} & 16.24\\
                     
            T5          & 16.45 & 25.84 & \textbf{29.88}
                        & 12.17 & 19.60 & 16.21\\
        \midrule
        &&&k = 1000\\
        \midrule
            DPR Reader  & 16.51 & 27.88 & 29.69 
                        & 15.10 & \textbf{20.83} & 16.24\\
                     
            T5          & 16.45 & 32.92 & \textbf{29.88}
                        & 12.17 & 19.60 & 16.21\\
        \bottomrule
    \end{tabular}
    \caption{
	    Multiple Answer Prediction Performance for Entity Questions. In extrinsic evaluation, BM25 is used to retrieve tweets. $k$ refers to, for each question, the number of tweets that are retrieved as context. 
	    $m$ is a hyperparameter that indicates the number of answers the system needs to predict.
	}\label{tab:MultiAnswerPrediction_entity}
\end{table}

\begin{table}[t]
    \centering
    \setlength{\tabcolsep}{8pt} 
    \renewcommand{\arraystretch}{1.2} 

    \begin{tabular}{lcccc}
        \toprule
            & \multicolumn{2}{c}{\textbf{Intrinsic}} & \multicolumn{2}{c}{\textbf{Extrinsic(E2E)}}\\\cmidrule(lr){2-3}\cmidrule(lr){4-5}
            \textbf{k} & 
            \Fanswer & \FcontroOne &
            \Fanswer & \FcontroOne\\
        \midrule
            5       & 89.71 & 54.58
                    & 78.02 & 17.77\\
                    
            10      & 91.83 & 50.98
                    & 91.09 & 19.04\\
        
            100     & 91.99 & 49.80
                    & 99.67 & 2.70\\
                     
            500     & 91.99 & 49.80
                    & 100   & 0.49\\
                    
            1000    & 91.99 & 49.80
                    & 100   & 0.25\\
        \bottomrule
    \end{tabular}
    \caption{
	    Multiple Answer Prediction Performance for Yes/No Questions. We use the best-performed tweet retriever and stance detection model in our previous experiments, i.e., BM25 + DeBERTa-large. $k$ refers to, for each claim, the number of tweets that are retrieved for stance detection.
	    E2E refers to end-to-end.
	}\label{tab:MultiAnswerPrediction_yesno}
\end{table}

\noindent\textbf{Results}:~~~~ Table~\ref{tab:MultiAnswerPrediction_entity} and Table~\ref{tab:MultiAnswerPrediction_yesno} show the performance on multiple answer prediction for entity questions and yes/no questions, respectively. As Table~\ref{tab:MultiAnswerPrediction_entity} presents, the pretrained DPR Reader (on NQ) performs marginally better than the pretrained T5 (on SQuAD) in extrinsic evaluation. Nonetheless, neither of the pretrained models perform well on this task, indicating there is huge room for improvement in multi-answer open-domain QA on social media. Promising next steps include (1) domain adaptive training over social media data; (2) using better answer normalization and selection rules to distinguish distinct answers from the same answers in different presentations even from the ones with typos which is quite common in social media, e.g., 'the USA', 'the United States', 'America', etc. The difference between the intrinsic and extrinsic evaluation scores shows how much the performance of the tweet retriever affects the final results.

Table~\ref{tab:MultiAnswerPrediction_yesno} shows how retrieval number($k$) affects the multiple answer prediction for yes/no questions. In this experiment, we use the best-performed tweet retriever and stance detection model in our previous experiments, i.e., BM25 + DeBERTa-large. We can see that, for intrinsic evaluation, the \Fanswer does not change much when $k$ is greater than 10. This is because, on average, there is only 5.4 relevant stance evidence annotated for each yes/no question. In extrinsic evaluation, we observe that, when $k$ is increased to 100, the \Fanswer of the model increases to close to 100, while \FcontroOne drops significantly to 2.70. 

Several insights can be drawn from this: (1) Unlike multi-answer predictions for entity questions, \Fanswer itself is insufficient to reflect the actual performance of the multi-answer prediction task for yes/no questions. The reason is that no matter how poor the retrieval or stance detector's performance may be, \Fanswer will always be close to 100 when the size of the retrieval is large enough. (2) However, the \Fcontro scores are indicative of the actual performance of the multi-answer prediction for the yes/no questions, since they compare the predicted evidence with the gold evidence. (3) In addition, we observe that the larger the retrieval number the worse the \FcontroOne score of the multi-answer prediction for the yes/no question. This reflects how the performance of the stance detection model detracts from the overall performance of the multi-answer question predictions because the top evidence is selected based on the logits of the stance detection model.  (4) Compared to the intrinsic setting, the drop in extrinsic performance illustrates the challenges within the tweet retrieval model.

\begin{table}[t]
    \centering
    \setlength{\tabcolsep}{3pt} 
    \renewcommand{\arraystretch}{1.2} 
    
    \begin{tabular}{l|c|ccc|c|ccc}
        \toprule
            & \multicolumn{4}{c|}{\textbf{Intrinsic}} & \multicolumn{4}{c}{\textbf{Extrinsic(E2E)}}\\
            
            \cmidrule(l{0em}r{0em}){2-5}\cmidrule(l{0em}r{0em}){6-9}
            
            & \Fanswer & \multicolumn{3}{c|}{\Fcontro} & \Fanswer & \multicolumn{3}{c}{\Fcontro}\\
            
            \cmidrule(l{0em}r{0em}){3-5}\cmidrule(l{0em}r{0em}){7-9}
            
            \textbf{Q Type}  &  &  e=1 & e=10 & e=100 & & e=1 & e=10 & e=100\\
            
        \midrule
            Entity      & 100   & 5.37  & 22.84 & 58.90 
                        & 20.83 & 0.71  & 4.34  & \textbf{12.48}\\
                     
            Yes/No      & 91.99 & 49.80 & 64.30 & 64.30 
                        & 100   & 0.25  & 4.78  & \textbf{29.04}\\
        \bottomrule
    \end{tabular}
    \caption{
	    Stance Mining and End-to-end MythQA Performance. Experiment setup: Retriever: BM25, retrieval number: 1000; machine reader: DPR Reader(on NQ), number of predicted answers: 5; stance detector: DeBERTa-large(on MNLI). Q refers to the question. $e$ refers to the number of stance(supporting/refuting) evidence included in the prediction. E2E refers to end-to-end.
	}\label{tab:StanceMining}
    
\end{table}

\subsection{Contradictory Stance Mining}\label{subsec: contradictory Stance Mining}

Contradictory stance mining aims to find contradictory(both supporting and refuting) claims for a given claim as described in section~\ref{sec: task_setup}.

For entity questions, we perform both intrinsic and extrinsic evaluations.

\noindent\textbf{Intrinsic}: Input claims are composed of gold answers and corresponding questions.

\noindent\textbf{Extrinsic}: Input claims are composed of the extrinsic MRC outputs and the corresponding questions. This is equivalent to an end-to-end MythQA evaluation.

For yes/questions, the experiment setting of contradictory stance mining is equivalent to multiple answers prediction as described in section~\ref{subsec: Multiple Answers Prediction}. \\

\noindent\textbf{Results}:~~~~ 
The intrinsic and extrinsic (end-to-end MythQA) stance mining results are presented in Table~\ref{tab:StanceMining}. In comparison to intrinsic evaluation results, the huge performance drop in the extrinsic evaluation of entity questions indicates that multiple-answer prediction is the bottleneck of the entire pipeline. For yes/no questions, the analysis presented in section~\ref{subsec: Multiple Answers Prediction} is also applicable since, as noted, in this setting, multi-answer prediction for yes/no questions is equivalent to contradictory stance mining and follows the same metrics.

Overall performance is low both intrinsically and extrinsically, which highlights the difficulty of the task. This can be attributed to two factors. First, multiple modules are involved in the pipeline. Mistakes accumulate at each stage. Further improvements can be made to each module. Second, there is a lack of annotated data, especially for open-domain QA over social media; future work can explore how to maximize the use of supervision from other data collected from social media.

\section{Conclusion \& Future Work}
We introduced MythQA, a new multi-answer open-domain question answering(QA) task that involves contradictory stance mining for query-based large-scale check-worthy claim detection.  We constructed TweetMythQA, a dataset with 5.3K contradictory evidence annotations on 1.2K distinct answers to 522 manually generated multi-answer questions. Furthermore, we present a baseline system for MythQA and evaluate existing NLP models for each component using the TweetMythQA dataset. We also highlight the potential areas for improvement.

Future research on MythQA may include (1) domain-adaptive training over social media data, (2) improved answer normalization and selection rules to distinguish distinct answers from the same answers in different presentations, including those with typos often seen in social media, and (3) more carefully evaluating its effectiveness in downstream fact-checking systems.





\begin{acks}
Our sincere thanks go out to the anonymous reviewers who took the time to provide constructive comments. This work is partially supported by DARPA under Award No. FA8750-18-2-0014 (AIDA/GAIA).
\end{acks}




\bibliographystyle{ACM-Reference-Format}
\balance
\bibliography{anthology,custom}

\end{document}